# Chart-RL: Policy Optimization Reinforcement Learning for Enhanced Visual Reasoning in Chart Question Answering with Vision Language Models

Yunfei Bai, Amit Dhanda, Shekhar Jain
Amazon
byunfei@amazon.com, amdhanda@amazon.com, shekhajt@amazon.com

## Abstract

The recent advancements in Vision Language Models (VLMs) have demonstrated progress toward true intelligence requiring robust reasoning capabilities. Beyond pattern recognition, linguistic reasoning must integrate with visual comprehension, particularly for Chart Question Answering (CQA) tasks involving complex data visualizations. Current VLMs face significant limitations in CQA, including imprecise numerical extraction, difficulty interpreting implicit visual relationships, and inadequate attention mechanisms for capturing spatial relationships in charts. In this work, we address these challenges by presenting Chart-RL, a novel reinforcement learning framework that enhances VLMs' chart understanding through feedback-driven policy optimization of visual perception and logical inference. Our key innovation includes a comprehensive framework integrating Reinforcement Learning (RL) from Policy Optimization techniques along with adaptive reward functions, that demonstrates superior performance compared to baseline foundation models and competitive results against larger state-of-the-art architectures. We also integrated Parameter-Efficient Fine-Tuning through Low-Rank Adaptation (LoRA) in the RL framework that only requires single GPU configurations while preserving performance integrity. We conducted extensive benchmarking across open-source, proprietary, and state-of-the-art closed-source models utilizing the ChartQAPro dataset. The RL fine-tuned Qwen3-VL-4B-Instruct model achieved an answer accuracy of 0.634, surpassing the 0.580 accuracy of the Qwen3-VL-8B-Instruct foundation model despite utilizing half the parameter count, while simultaneously reducing inference latency from 31 seconds to 9 seconds. We also performed comprehensive comparative analysis of Chain of Thought (CoT) reasoning performance across those models, demonstrating that the Chart-RL framework significantly enhanced the visual reasoning.

## 1 Introduction

The recent advancements in Large Language Models (LLMs) have been towards demonstrating true intelligence that requires more than pattern recognition. True intelligence demands robust reasoning capabilities, which involve processing existing knowledge through logical deduction to arrive at novel conclusions [23]. The reasoning capabilities are essential for complex, multi-step tasks such as mathematical problem solving, logical inference, and scientific analysis. The reasoning challenges are magnified in the domain of VLMs, where linguistic reasoning must be fused with visual comprehension. Visual reasoning for VLMs involves two interconnected steps: accurately recognizing key information from visual inputs and conducting complex reasoning based on that extracted data. Data visualizations, particularly charts, are high-density carriers of complex information. Interpreting them is not a trivial task as it necessitates a sophisticated sequence of visual reasoning capabilities that combine chart reading (recognition) and chart understanding (comprehension) to decipher underlying data patterns. Thereby, Chart Question Answering (CQA) has emerged as the standard, rigorous task to probe the visual reasoning capabilities of VLMs.

There are challenges in VLMs when handling CQA tasks, including less precise numerical extraction and computation for complex visual elements such as overlapping lines or stacked bars, implicit relationships between visual elements such as trends over time or proportional relationships, and the context-dependent nature of chart interpretation. This is due to the attention mechanisms in the VLM foundation model may not be able to effectively capture spatial relationships and hierarchical structures in charts, while the sequential processing of visual information does not align well with the non-linear nature of chart interpretation. Furthermore, for complex multi-step reasoning questions that require correlation of information from different parts of the chart or performing comparative analyses across multiple data points, VLMs often lack the ability to maintain and update intermediate results, along with the combination of numerical computation and visual understanding.

To address these challenges, in this paper, we present Chart-RL, a reinforcement learning framework to enhance VLMs' chart understanding and reasoning capability by creating a feedback-driven learning loop to optimize visual perception and logical inference. Our approach allows the model to learn complex visual-textual relationships inherent in charts, including data trends, correlations, and contextual information, while simultaneously improving its ability to generate coherent explanations and draw logical conclusions from visual data representations. It enables the VLMs to explore different reasoning pathways and reinforces successful strategies for chart analysis, leading to more robust performance on tasks requiring both visual comprehension and analytical reasoning. Our key contributions include

- We implemented Chart-RL with three advanced policy optimization techniques: Group-based Reinforcement Learning from Policy Optimization (GRPO), Direct Advantage Policy Optimization (DAPO), and Group Sequence Policy Optimization (GSPO). This comprehensive framework demonstrates enhanced performance in complex chart reasoning tasks, achieving superior accuracy compared to baseline foundation models and competitive performance relative to larger state-of-the-art architectures.
- Our reinforcement learning methodology incorporated Parameter Efficient Fine-Tuning (PEFT) through Low-Rank Adaptation (LoRA), which significantly reduces computational overhead and training costs while preserving model performance integrity. This approach enables efficient model optimization without compromising the quality of understanding and reasoning capabilities of the chart.
- We conducted benchmarking across open-source, proprietary, and state-of-the-art closed-source models utilizing the ChartQAPro dataset, demonstrating the CQA accuracy improvement and latency reduction.
- We performed a comprehensive comparative analysis of the performance of chart reasoning in open-source foundation models, proprietary foundation models, and state-of-the-art closed-source models, establishing the efficacy of policy optimization-based reinforcement learning to improve model performance.

## 2 Related Work

### 2.1 Chart question answering benchmarks

CQA tasks have evolved from synthetic, template-driven tasks toward more diverse settings. Earlier datasets for CQA, such as FigureQA [20] and DVQA [19], were mostly generated with synthetic data using a plotting software, where the questions are generated using a small number of templates and the answers from a fixed set of vocabulary, such as 'yes' and 'no'. PlotQA [9] expanded coverage to scientific plots and emphasized real-valued answers. More recent benchmarks aim to better reflect real-world chart diversity and question styles. ChartX/ChartVLM [25] dataset was created by synthesizing data and questions using GPT-4 to generate multiple chart types and a broader suite of chart reasoning tasks, including summarization and redrawing, alongside a foundation model for complex chart reasoning. The questions from these synthetic datasets are more template-based than those that require visual reasoning. Complementary efforts such as CharXiv [24] highlighted gaps in multimodal LLM chart understanding on figures drawn from scientific papers. It relied on charts sourced exclusively from arXiv papers, however with limited visual and topical diversity and lacking numerous real-world question types. ChartMuseum [11] provides targeted evaluations of the reasoning of the LVLM chart and helps characterize persistent failure modes beyond surface-level extraction. On the other hand, ChartQA [21] included real-world data from sources like Our World in Data, Statista, OECD, and Pew Research, and includes both human and machine-authored questions to provide a diverse spectrum of question styles. However, the chart types are limited to line, bar, and pie charts with largely factoid questions that require simple data extraction or basic arithmetic. ChartQAPro [22] further increased difficulty and diversity. It included real-world charts sourced from diverse online platforms with a wide variety of human-verified question-answer pairs that are factoid, multiple-choice, hypothetical, conversational, multi-chart, and unanswerable cases.

### 2.2 From chart reading to chart understanding

Chart understanding is closely related to document and infographic VQA, where OCR, layout, and grounded language understanding interact. DocVQA [15] focused on questions over document images with strong dependence on textual content and spatial arrangement. InfographicVQA [16] expanded to visually rich layouts that combine text blocks, icons, and charts. Pix2Struct [17] suggested that large-scale pretraining on screenshot parsing can provide useful inductive biases for these structured, text-heavy visual domains. Many failures in CQA originate in the perception stage: extracting accurate text, marks, and values under clutter, overlaps, and diverse encodings. ChartOCR [14] utilized chart-specific pipelines that combine detection, OCR, and structural parsing, where the extraction errors can bottleneck downstream reasoning. Therefore, an increasingly common strategy is to disentangle perception and reasoning by converting charts into structured representations: DePlot [12] translated charts into tables that can be consumed by a language model for subsequent logical and numerical reasoning. Building on this direction, SIMPLOT [13] improved CQA by distilling essential chart information and reducing irrelevant visual content, which is especially helpful for multi-step comparisons and aggregation.

### 2.3 Multimodal understanding and reasoning in LLMs and VLMs

Progress in CQA is tightly coupled to advances in general-purpose multimodal LLMs (MLLMs). Proprietary frontier models, such as Anthropic Claude models [27, 28], Google Gemini models [30], and OpenAI GPT models[33] often provide strong zero-shot performance, however are difficult to customize due to limited transparency. In parallel, open-source VLMs such as InternVL [32], Janus [29], and Qwen-VL [26, 31], combine LLM with vision encoders to understand, interpret, and relate visual data, meanwhile enabling controlled adaptation, including targeted prompting, domain-specific tuning, and architectural ablation. This is critical to improve chart-specific visual understanding and reasoning through multi-step reasoning from MLLMs and VLMs at inference time. Chain-of-Thought (CoT) prompting [1] can improve performance on tasks requiring multi-hop deduction and arithmetic. ReAct [2] extends CoT by interleaving reasoning with actions to update intermediate state and recover from errors. To improve CoT reasoning, Reinforcement Learning with Verifiable Rewards (RLVR) [3] is an effective approach that incentivizes correct and logical thought chains through rewards in verifiable domains. Among the popular RLVR methods, Group Relative Policy Optimization (GRPO) [4] captures the inherent variability in reasoning processes and trains the model to consistently outperform its own average performance. Building upon the success of GRPO, Direct Advantage Policy Optimization (DAPO) [5] employs a higher clip ratio to encourage more diverse and exploratory thinking processes for more efficient training process. Group Sequence Policy Optimization (GSPO) [6] shifts the optimization to the sequence level for more sophisticated reasoning and computational efficiency improvements.

## 3 Methodology

We introduce a VLM reinforcement learning framework comprising three distinct pipelines, as shown in Figure 1. The first constitutes a data pre-processing pipeline for training and inference data preparation. The second is an inference pipeline that facilitates chart understanding and reasoning tasks, formally defined as follows. Given a chart image $I$ and a corresponding textual question $Q$, the objective function $F$ seeks to derive a textual and/or numerical answer $A$ to the question $Q$, accompanied by a Chain of Thought reasoning procedure $R$, where $A \in R$. This relationship is expressed as:

$$F : (I, Q) \rightarrow (A, R) \quad (1)$$

The third pipeline encompasses the fine-tuning methodology for VLM chart reasoning optimization and generalization. While traditional approaches employ an initial stage to establish foundational reasoning paradigms through Supervised Fine-Tuning (SFT) using human-labeled or large-scale state-of-the-art MLLM-distilled datasets $(A, R)$, our methodology directly applies fine-tuning to foundation models without requiring preliminary SFT stages. Consequently, our training pipeline in Figure 1 employs policy optimization-based reinforcement learning to refine VLM chart interpretation capabilities and improve the accuracy and reliability of the answers for CQA tasks.

Our training pipeline incorporates GRPO, DAPO and GSPO, which eliminate the requirement for value functions. This approach enables VLMs to develop autonomous reasoning

---

**Algorithm 1**

**Require:** chart image $I$, textual question $Q$, initial policy model $W$, reward function Rew

Pre-process the image to reduce the image size down to fixed width of 300 pixels
**for** step = $1 \dots M$ **do**
  Sample a batch $(I_b, Q_b)$ from $(I, Q)$
  Create prompt $D$
  Infer the model policy to get a group of $G$ candidate responses $\{A_1, A_2, \dots, A_G\}$
  Calculate reward for each candidate, to generate $\{\text{rew}(A_1), \text{rew}(A_2), \dots, \text{rew}(A_G)\}$
  Compute relative advantage $a_i$
  **for** iteration = $1, \dots, q$ **do**
    Update policy model $W$ by adjusting model parameters to increase positive advantage
  **end for**
**end for**

**Ensure:** Updated policy model $W$

---

capabilities through iterative policy refinement based on reward signals derived from chart interpretation accuracy.

Formally, for each input training sample $(I, Q)$, we infer the model policy to obtain a group of $G$ candidate responses $\{A_1, A_2, \dots, A_G\}$. Each answer $A_i$ receives a reward $\text{rew}(A_i)$, and these rewards are utilized to compute a normalized relative advantage $a_i$ for each sample:

$$a_i = \frac{\text{rew}_i - \mu(\text{rew}_i, \dots, \text{rew}_G)}{\sigma(\text{rew}_i, \dots, \text{rew}_G)} \quad (2)$$

Subsequently, the VLM parameters are updated to increase the probability of actions with positive advantages, while a Kullback-Leibler (KL) divergence penalty against the foundation model used as reference model, which ensures training stability.

### 3.1 Image Pre-processing

To optimize computational efficiency while maintain visual fidelity, we implemented an image preprocessing pipeline that standardizes the chart dimensions. The original chart images were resized to maintain a fixed width of 300 pixels while preserving the aspect ratio, resulting in proportionally scaled heights. This reduction significantly decreased the memory footprint and processing time compared to using full-resolution images (average original size greater than 1000 pixels in width) without compromising the model's ability to extract relevant visual information. We validated this approach by comparing model performance between reduced and full-resolution images, finding no significant degradation in accuracy while achieving a 70% reduction in memory usage and inference time.

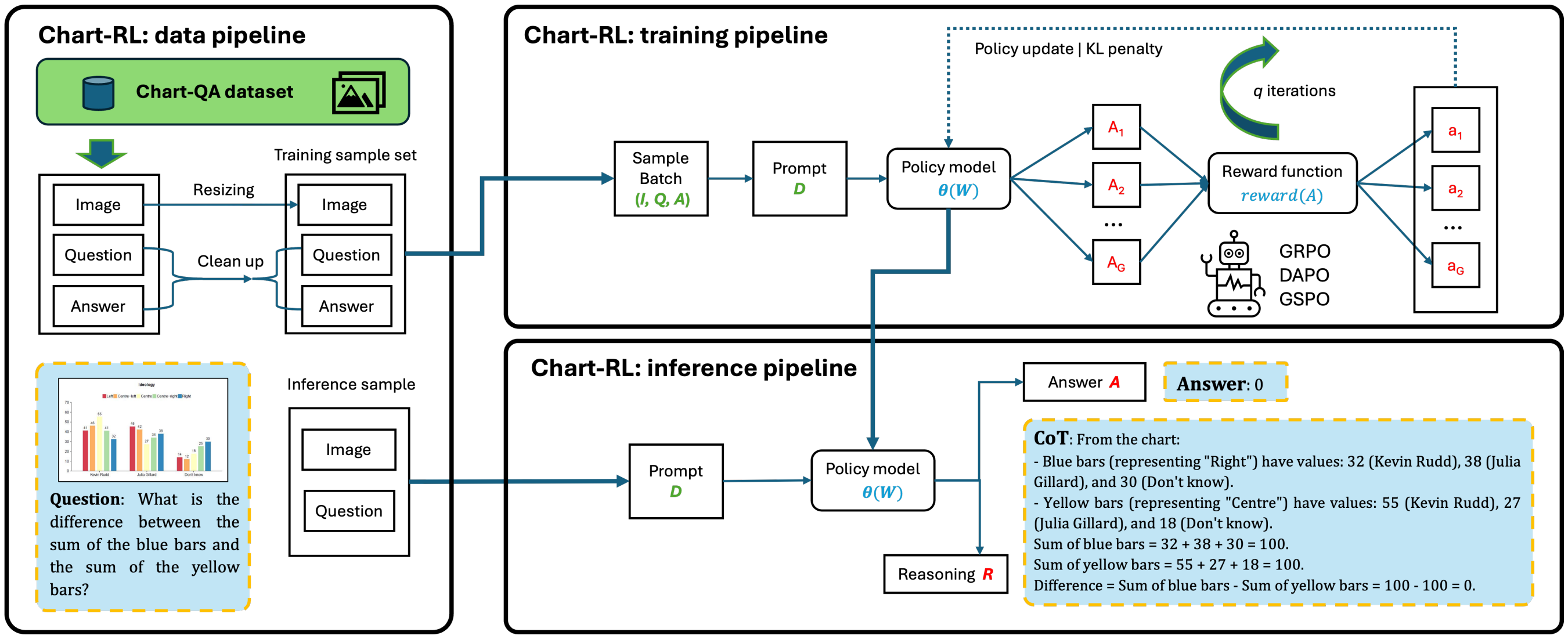

**Figure 1.** Overview of Chart-RL Framework. We first use data pipeline to construct training samples from pre-processed CQA dataset, then factorize policy optimization based reinforcement learning (GRPO/DAPO/GSPO) in the training pipeline. The trained policy model is then used in the inference pipeline to complete question-answering task through the enhanced visual understanding and reasoning by the VLMs.

### 3.2 Policy Optimization Techniques

We employed three RLVR methods in our framework, users can select the most appropriate method based on their tasks.

- GRPO generates groups of responses and evaluates each against the average score of the group, rewarding those performing above average while penalizing those below. This relative comparison mechanism creates a competitive dynamic that encourages the model to produce higher-quality reasoning.
- DAPO employs approximately 30% higher clip ratio than GRPO to eliminate less meaningful samples and improve overall training efficiency, applies token-level policy gradient loss to provide more granular feedback on lengthy reasoning chains, and incorporates overlong reward shaping to discourage excessively verbose responses that waste computational resources.
- GSPO utilizes the sequence-level importance weights rather than the token-level weights in GRPO, to avoid high variance and unstable gradients especially for the long text outputs in the Mixture-of-Experts (MoE) model training.

### 3.3 Reward Function Design

For Reinforcement Learning, we design the reward function $\text{Rew}(A_i)$ as a composite of three types of reward, format reward for answer format, accuracy reward for answer accuracy, and reasoning reward for reasoning in response.

$$\begin{aligned} \text{rew}(A_i) = \text{rew}_{\text{format}}(A_i) + \text{rew}_{\text{accuracy}}(A_i) \\ + \text{rew}_{\text{reasoning}}(A_i) \end{aligned} \tag{3}$$

The format reward function $\text{rew}_{\text{format}}(A_i)$ implements a binary scoring mechanism to promote interpretability of the response and structural consistency. Responses that comply with the established template—featuring CoT reasoning encapsulated within <think> </think> XML tags and answers delimited by <answer> </answer> XML tags—receive a positive reward of “1”. Non-compliant responses that deviate from this formatting specification in either the reasoning or answer components are penalized with a reward of “0”.

For evaluation of the accuracy $\text{rew}_{\text{accuracy}}(A_i)$ and reasoning $\text{rew}_{\text{reasoning}}(A_i)$ rewards, we implement a unified Large Language Model-based reward mechanism. This reward function employs OpenAI’s GPT-4 model (gpt-oss-120B), which has superior reasoning capabilities that render it suitable for serving as an automated evaluator. We instruct the model to assess both the correctness of the predicted answer and the validity of the reasoning process leading to that answer. The reward function employs a three-tier categorical scoring system: a reward of “1” is assigned when the predicted answer is correct AND the reasoning process can logically lead to the correct answer; a reward of “0.5” is assigned when the predicted answer is correct BUT the reasoning process cannot logically support the correct answer; and a reward of “0” is assigned when the predicted answer is incorrect, regardless of the reasoning quality.

It is important to note that although our LLM-based reward mechanism utilizes another LLM, it is fundamentally different from an unsupervised critic. The reward function receives the ground truth answer from the ChartQAPro dataset and evaluates the predicted response against this known correct answer. The LLM judge serves as a sophisticated semantic matching function and reasoning verifier - assessing whether the predicted answer is equivalent to the ground truth while tolerating surface-level variations in phrasing, formatting, and numerical precision (e.g., '5,200' vs '5200' or '1.289' vs '1.3'). It further evaluates the model's ability to generate valid reasoning for query resolution. This is analogous to how the environment provides ground truth signals in traditional RL; here the dataset's human-verified answers constitute the environment's ground truth, and the LLM judge operationalizes the comparison and validation.

### 3.4 PEFT/LoRA Adoption

Within our Chart-RL framework, we adopt LoRA training representing a paradigmatic shift toward computationally efficient model optimization, and addressing the resource constraints inherent in traditional full weight training approach. In our policy optimization learning process, we employ rank decomposition matrices to update only a small subset of model parameters, reducing trainable parameters to less than 0.5% of the original model parameter size while maintaining the frozen base model weights.

This configuration reduces susceptibility to catastrophic forgetting problems common in traditional fine-tuning approaches, also enhances training efficacy by enabling efficient task switching through small, swappable matrices that require minimal storage compared to full model copies. Meanwhile, this design enables our RL implementation on single GPU configurations rather than requiring distributed computing infrastructure, resulting in rapid experimentation and deployment cycles, making reinforcement learning optimization more accessible for production environments while maintaining performance integrity comparable to resource-intensive full fine-tuning methodologies.

## 4 Experiments

We conducted a comprehensive experimental evaluation across multiple MLLMs and VLMs, including RL fine-tuned variants of Qwen3-VL-4B-Instruct trained via GRPO, DAPO, and GSPO methodologies. The ChartQAPro dataset was used to assess chart image understanding and question-answering capabilities. All model training and inference procedures were executed on a single GPU compute node with 24GB memory, demonstrating computational efficiency. Model performance was benchmarked using two primary metrics: answer accuracy and inference latency, providing a comprehensive assessment of both predictive quality and computational efficiency. The answer-accuracy is measured by an LLM-as-a-judge (OpenAI gpt-oss-120b) that compares the ground truth answers in ChartQAPro with the model inferenced answers.

### 4.1 Data Preparation

The ChartQAPro dataset, which contains diverse real-world charts with human-verified question-answer pairs. The dataset consists of 1,948 samples, with the last 500 samples (excluding unanswerable questions) used as a test set and the remaining samples for training. The question types are distributed across five categories: factoid questions (55%, n=1,081) that focus on direct data extraction and simple calculations, conversational questions (16%, n=311) that require contextual understanding and follow-up reasoning, fact-checking questions (13%, n=244) that test verification and validation capabilities, multiple-choice questions (11%, n=214) that evaluate discriminative reasoning, and hypothetical questions (5%, n=98) that assess predictive and extrapolative reasoning. The charts are sourced from diverse online platforms, covering various visualization types including bar charts, line charts, pie charts, and scatter plots. We employed the image preprocessing pipeline to resize the dataset images, thereby ensuring compatibility with GPU memory requirements for the GRPO, DAPO, and GSPO training procedures in the experiments.

### 4.2 Policy Optimization

For training implementation, we configured LoRA adapters with rank 256 and alpha value of 1024 for parameter-efficient fine-tuning (PEFT), targeting the query and value projection layers with a dropout rate of 0.05. The GRPO training was conducted with a learning rate of 1e-5 and a batch size of 2 per device, using bf16 precision. The model was trained for one epoch with maximum prompt and completion lengths of 8192 and 1024 tokens respectively, generating 2 responses per input for group comparison. The accuracy reward function (as detailed in Section 3.2) was used to evaluate and optimize the model's chart reasoning capabilities.

The training processes for GRPO, DAPO and GSPO on Qwen3-VL-4B-Instruct, including training loss, reward, entropy and completion mean length, are depicted in Figure 2. For all three optimization approaches, throughout the training process, the reward trajectory exhibited a consistent upward trend, while the entropy values demonstrated a monotonic decrease, indicating a progressive stabilization of the learning dynamics. These dual indicators collectively confirm a successful convergence towards the specified optimization objective. In particular, the completion mean length stabilized after approximately 150 training steps, demonstrating rapid convergence and efficient attainment of the training objectives. This shows successful policy optimization with consistent refinement of the model's chart reasoning capabilities.

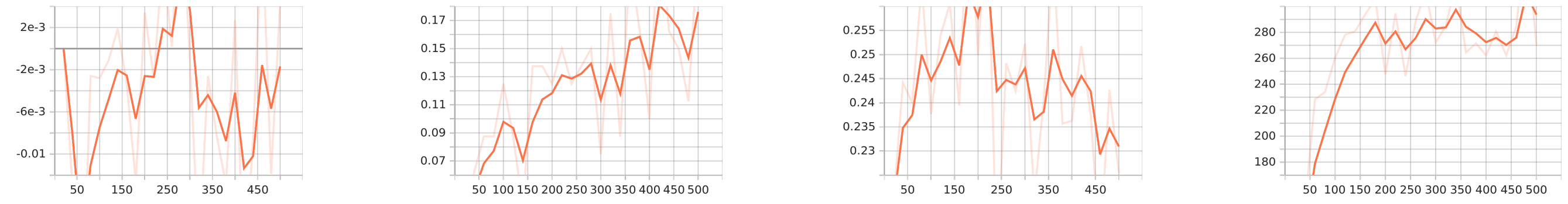

GRPO: Training loss | GRPO: Reward | GRPO: Entropy | GRPO: Mean length

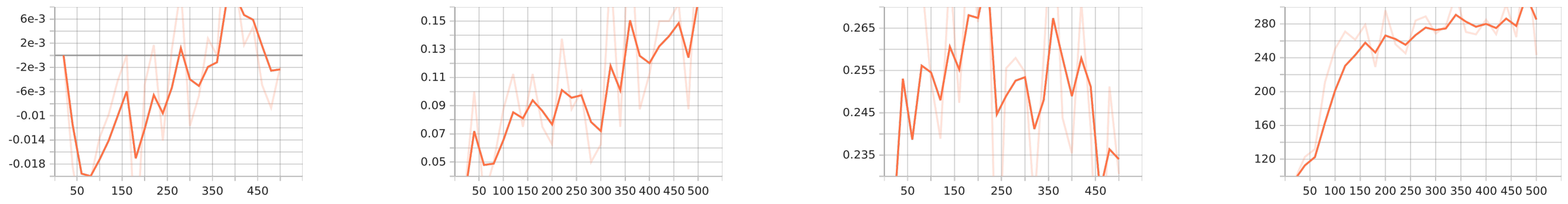

DAPO: Training loss | DAPO: Reward | DAPO: Entropy | DAPO: Mean length

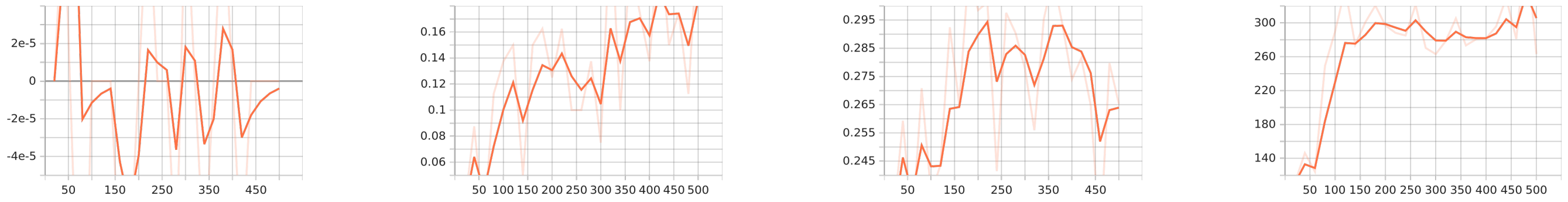

GSPO: Training loss | GSPO: Reward | GSPO: Entropy | GSPO: Mean length

**Figure 2.** Reinforcement Learning Processes by GRPO, DAPO and GSPO. The training loss represents model's error decreasing as it learns from the training data. The RL reward indicates the upward trend over time during the successful training run. The policy entropy demonstrates a stable non-zero level indicating the active, flexible and exploring training process. The episode mean length shows rapid and stable convergence over the reinforcement learning.

### 4.3 Quantitative Analysis on CQA

The experimental results demonstrate significant improvements achieved through our reinforcement learning framework, as shown in Table 1. In our experimental evaluation, we benchmarked the question-answer accuracy of RL fine-tuned models against two distinct baseline categories. The first baseline comprises state-of-the-art (SOTA) MLLMs (specifically, Claude Sonnet 3.7 and Claude Sonnet 4.5) representing general-purpose architectures capable of processing heterogeneous data modalities, including both textual and visual inputs. Claude Sonnet 3.7 achieved an accuracy of 0.769 on the evaluation dataset. The second baseline encompasses a collection of popular open-source vision-language models (VLMs) with relatively compact architectures (typically 7-8B parameters), including Qwen2-VL, Qwen3-VL, Janus-Pro, InternVL, and LLaVA. Among these models, Qwen3-VL-8B-Instruct demonstrated superior performance with an accuracy of 0.580 and a mean inference latency of 31.59 seconds.

In a vertical comparison of model performance, reinforcement learning fine-tuning substantially improved the accuracy of the Qwen3-VL-4B-Instruct foundation model. Specifically, the GSPO-fine-tuned variant improved accuracy from baseline 0.396 to 0.622, while the GRPO and DAPO fine-tuned variants achieved further improvements, reaching accuracies of 0.627 and 0.634, respectively. Notably, despite utilizing only half the parameter count, all three fine-tuned 4B models surpassed the performance of the larger Qwen3-VL-8B-Instruct baseline model (accuracy: 0.580), demonstrating the efficacy of reinforcement learning fine-tuning in achieving parameter-efficient performance gains.

The fine-tuned models exhibited marked improvements in computational efficiency, achieving inference latencies between 9.48 and 9.69 seconds—a 71% reduction relative to the Qwen3-VL-8B-Instruct model (31.59 seconds). Although closed-source state-of-the-art models such as Claude Sonnet 3.7 retain an accuracy advantage (0.769 vs. 0.634 for our best model), our reinforcement learning fine-tuning methodology delivers competitive performance while substantially reducing resource requirements. Critically, these results were obtained using only a single GPU with 24GB memory, demonstrating exceptional parameter and hardware efficiency. The latency-accuracy analysis presented in Figure 3 illustrates that our models cluster within the Pareto-optimal region, simultaneously achieving high accuracy and minimal inference latency, a favorable trade-off for deployment in resource-constrained production environments. In the Figure, we present the performance distribution of evaluated models within a latency-accuracy coordinate space. This visualization facilitates model selection based on application-specific trade-offs between prediction accuracy and inference latency. You may prioritize models with marginally reduced accuracy but substantially lower latency, accepting minor performance degradation in exchange for optimized computational efficiency tailored to their deployment constraints. Particularly, the reinforcement learning fine-tuned Qwen3-VL-4B-Instruct models occupy the lower-right quadrant, demonstrating superior accuracy coupled with reduced inference latency relative to baseline foundation vision-language models, thus establishing a Pareto-efficient frontier for this task.

**Table 1.** Model Performance Comparison. The first set of models are the state-of-the-art (SOTA) proprietary (or closed-source) MLLMs, the second set are the open-source models, while the third set are our fine-tuned Qwen models

| **Model Name** | **Acc(0-1)** | **Latency(s)** |
|---|---|---|
| *SOTA MLLM* | | |
| Claude Sonnet 3.7 | 0.769 | – |
| Claude Sonnet 4.5 | 0.750 | – |
| *Open-source VLM* | | |
| Qwen2.5-VL-7B-Instruct | 0.511 | 6.72 |
| Qwen3-VL-4B-Instruct | 0.396 | 10.04 |
| Qwen3-VL-8B-Instruct | 0.580 | 31.59 |
| Janus-pro-7B | 0.320 | 18.25 |
| InternVL3.5-8B-MPO | 0.420 | 27.27 |
| Llava-v1.6 | 0.570 | 6.19 |
| *Fine-tuned VLM* | | |
| Qwen3-VL-4B-Instruct-GRPO | 0.627 | 9.84 |
| Qwen3-VL-4B-Instruct-DAPO | **0.634** | 9.48 |
| Qwen3-VL-4B-Instruct-GSPO | 0.622 | 9.69 |

## 5 Chart Image Understanding and Reasoning Analysis

The enhanced reasoning capabilities are evident in the improved accuracy scores. The GRPO, DAPO and GSPO variant models excelled in handling complex multi-step reasoning tasks, demonstrating more coherent Chain of Thought reasoning compared to baseline models. The following example analysis indicates that our policy optimization techniques effectively enhance both the accuracy of final answers and the quality of intermediate reasoning steps.

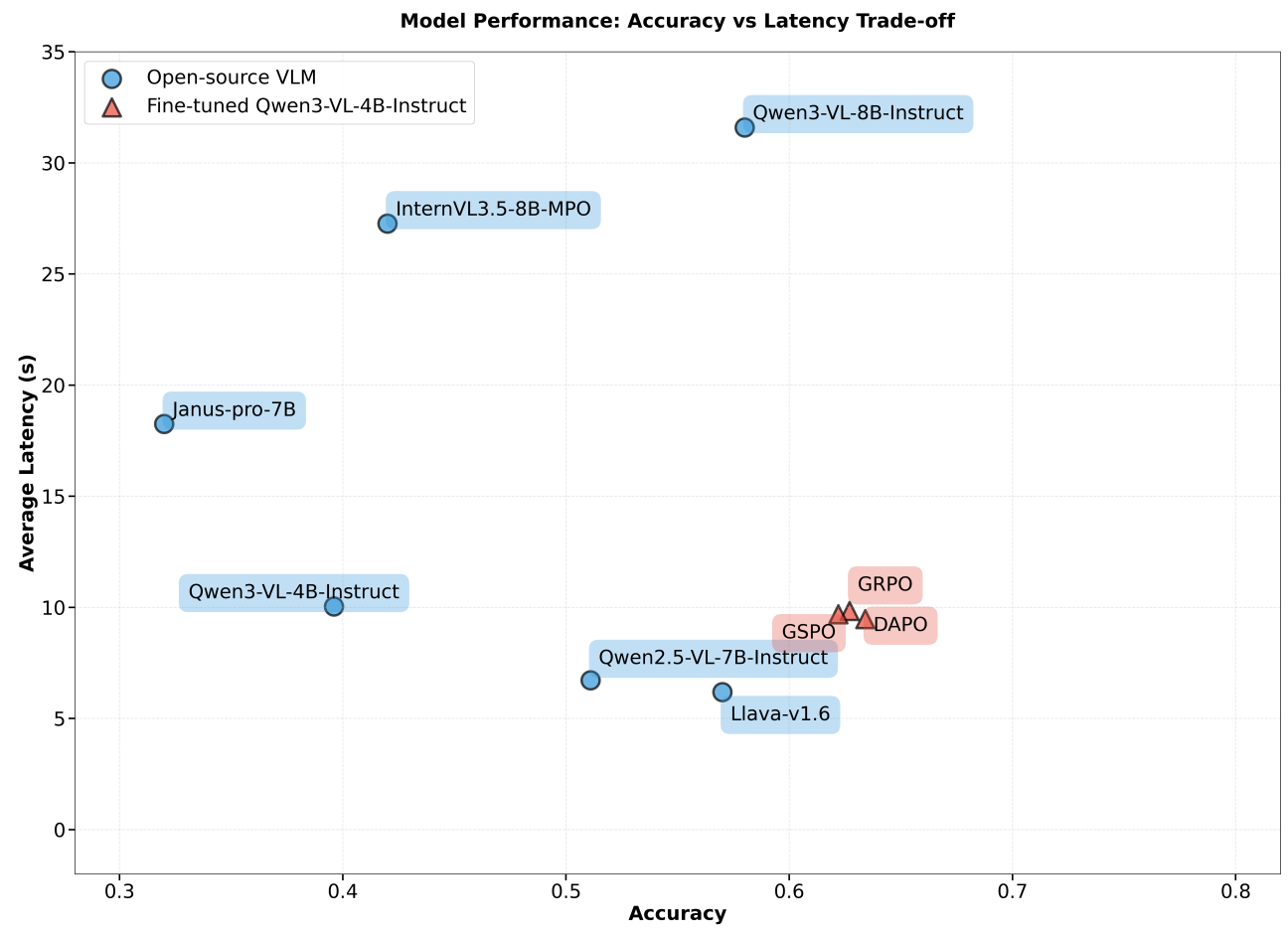

**Figure 3.** Latency-Accuracy trade-off analysis. Our RL-enhanced models achieve an optimal balance between qualitative performance and computational efficiency.

### 5.1 Example 1: Open-ended Question

**Question**: What is the difference between the sum of the blue bars and the sum of the yellow bars?

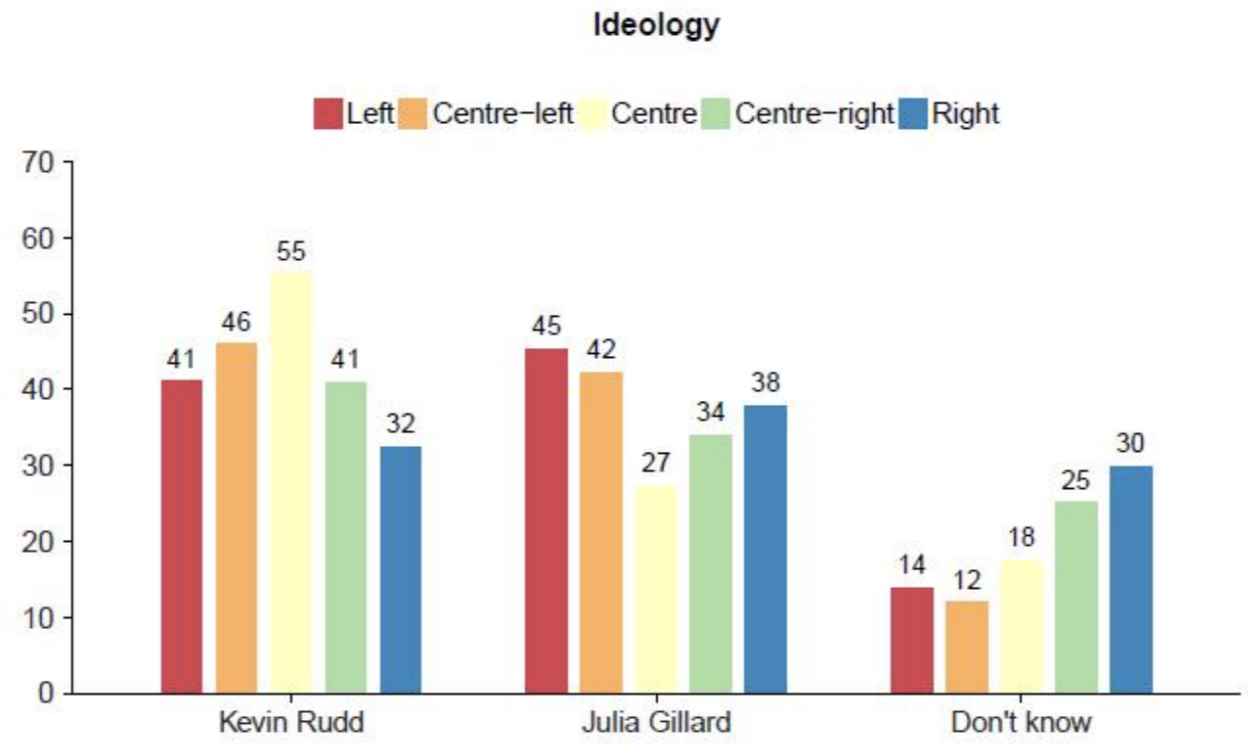

**Ground Truth**: 0

**Model Responses**:

- Claude 3.7 Sonnet: *The sum of the blue bars (Right) is 32 + 38 + 30 = 100. The sum of the yellow bars (Centre) is 55 + 27 + 12 = 94. The difference between the sum of the blue bars and the sum of the yellow bars is 100 - 94 = 6.*
- Qwen3-VL-4B-Instruct: *12*
- Qwen3-VL-4B-Instruct-GRPO: *0*

- Qwen3-VL-4B-Instruct-DAPO: *0*
- Qwen3-VL-4B–Instruct-GSPO: *0*

This example illustrates a multi-step reasoning task requiring the vision-language model to: (1) identify all blue and yellow bars with their corresponding numerical labels, (2) compute separate summations for each color category, and (3) perform a comparative analysis between the two aggregated values. Both Claude Sonnet 3.7 and the baseline Qwen3-VL-4B-Instruct model generated erroneous responses. Specifically, the Claude model exhibited a visual perception error, misidentifying a yellow bar from the "Don't know" category by selecting the bar with value 12 instead of the correct value of 18. This misclassification propagated through the computational chain, yielding an incorrect summation of 55 + 27 + 12 = 94 rather than the ground truth of 55 + 27 + 18 = 100. Although the baseline Qwen3-VL-4B-Instruct model similarly produced an incorrect response, the reinforcement learning fine-tuned variants demonstrated robust performance by accurately identifying both blue and yellow bar categories and correctly executing the requisite mathematical operations (summation and subtraction). These results demonstrate that reinforcement learning fine-tuning yields substantial improvements across two critical dimensions: visual understanding and object classification accuracy, and multi-step mathematical reasoning capability.

### 5.2 Example 2: Multiple Choice

**Question**: Assuming the growth trend from 2010 to 2014 continues, what would be the estimated pension fund assets in 2015 (in n billion)?
(a) 4,800 (b) 5,200 (c) 5,500 (d) 6,000

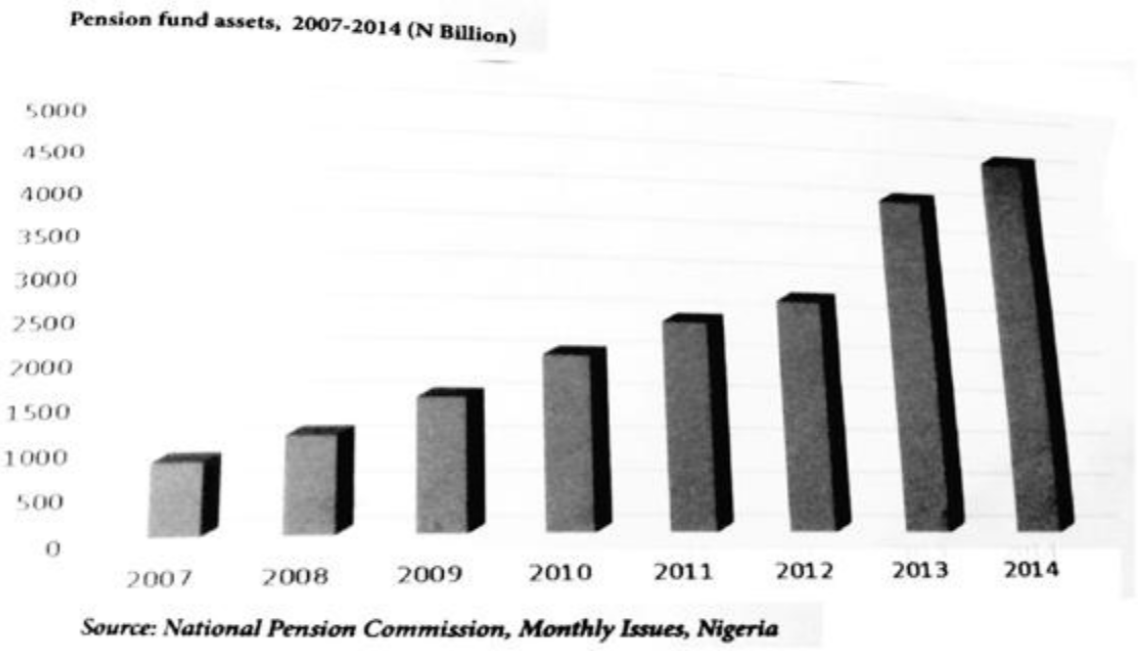

**Ground Truth**: 0

**Model Responses**:

- Claude 3.7 Sonnet: *Based on the chart showing pension fund assets from 2007-2014, I need to estimate the 2015 value by analyzing the growth trend from 2010-2014. Looking at the approximate values from the chart:*
  *- 2010: ~1,800 N billion*
  *- 2011: ~2,400 N billion*
  *- 2012: ~2,700 N billion*
  *- 2013: ~3,800 N billion*
  *- 2014: ~4,500 N billion*
  *To find the overall growth trend, I'll calculate the compound annual growth rate (CAGR) from 2010 to 2014:*
  *CAGR = (Final Value/Initial Value)^(1/number of years) - 1*
  *CAGR = (4500/1800)^(1/4) - 1*
  *CAGR = (2.5)^0.25 - 1*
  *CAGR = 1.26 - 1 = 0.26 or 26*
  *Applying this growth rate to estimate 2015: 2015 estimate = 4500 X (1 + 0.26) = 4500 X 1.26 = 5670 N billion*
  *This is closest to option c) 5,500 N billion. Therefore, the answer is c) 5,500 N billion.*
- Qwen3-VL-4B-Instruct: *uncertain*
- Qwen3-VL-4B-Instruct-GRPO: *b) 5,200*
- Qwen3-VL-4B-Instruct-DAPO: *b) 5,200*
- Qwen3-VL-4B–Instruct GSPO: *b) 5,200*

This task required the vision-language model to identify temporal patterns in pension fund asset values from 2010 to 2014 and construct an appropriate mathematical model to extrapolate the 2015 value. The Claude Sonnet 3.7 model employed only the endpoint values (2010 and 2014) to construct a Compound Annual Growth Rate (CAGR) model to estimate the 2015 value, demonstrating limited reasoning capability for this complex temporal prediction task. The baseline Qwen3-VL-4B-Instruct model failed to generate a valid response. In contrast, the reinforcement learning fine-tuned variants exhibited sophisticated multi-step reasoning: (1) extracting asset values across all intermediate years (2010-2014), (2) identifying the non-linear nature of year-over-year growth patterns, (3) selecting the most recent growth increment (2013-2014: 700) as the optimal predictor, (4) computing the projected 2015 value (4500 + 700 = 5200), and (5) correctly identifying answer option B. This performance demonstrates that reinforcement learning fine-tuning substantially enhances both visual data extraction accuracy and temporal reasoning capabilities in complex forecasting tasks.

### 5.3 Example 3: True/false Question

**Question**: The chart image in top left corner indicates a positive correlation between the account value of open cases and their creation date.

**Ground Truth**: TRUE

**Model Responses**:

- Claude 3.7 Sonnet: *The statement is incorrect. The chart in the top left corner titled "OPEN CASES: BY CREATED DATE" does not show a positive correlation between account value and creation date. Looking at the scatter plot, the account values (Y-axis ranging from* $0 to approximately 150K$*) appear to be distributed relatively randomly across the entire date range (X-axis*

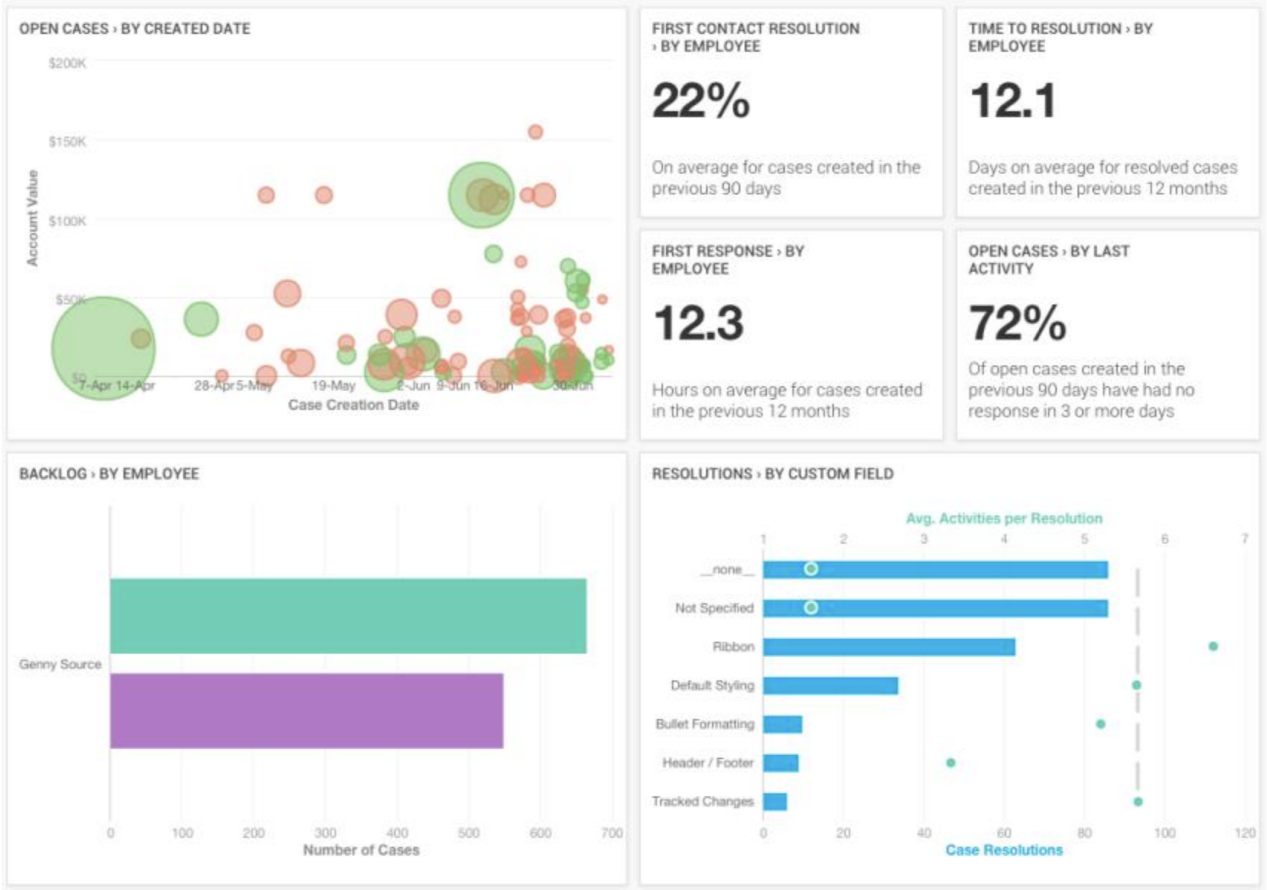

*from April to August). High-value accounts (around 100K – 150K) and low-value accounts are scattered throughout the time period without showing a clear upward trend from earlier to later dates. If there were a positive correlation, we would expect to see account values consistently increasing as we move from left (earlier dates) to right (later dates) on the chart, which is not evident in this visualization.*

- Qwen3-VL-4B-Instruct: *uncertain*
- Qwen3-VL-4B-Instruct-GRPO: *True*
- Qwen3-VL-4B-Instruct-DAPO: *True*
- Qwen3-VL-4B-Instruct GSPO: *True*

This task evaluates the model's capacity for statistical reasoning in visual contexts, requiring: (1) extraction of quantitative variables (account values, creation dates), (2) comprehension of correlation concepts, and (3) assessment of the relationship directionality. Claude Sonnet 3.7 exhibited conceptual confusion, mischaracterizing positive correlation as monotonic increase rather than recognizing the correct definition: a tendency for higher values of one variable to co-occur with higher values of another. Specifically, the model failed to identify that larger bubbles (higher account values) systematically appear at later temporal positions, while smaller bubbles (lower account values) concentrate at earlier dates. The baseline Qwen3-VL-4B-Instruct model demonstrated complete task failure. In contrast, the models enhanced by reinforcement learning accurately performed spatial pattern recognition, identifying the clustering of lower-value accounts ($50K) at earlier dates and higher-value accounts ($200K) at later dates, correctly inferring the positive correlation and selecting "True." This performance demonstrates enhanced capabilities in both visual data interpretation and statistical reasoning.

# 6 Conclusion

In this paper, we present a novel Chart-RL framework that significantly advances VLMs' capabilities in chart understanding and reasoning through innovative policy optimization techniques. Our comprehensive approach, integrating Group-based Reinforcement Learning from Policy Optimization techniques with adaptive reward functions using LLM judge, demonstrates substantial improvements in chart reasoning performance while maintaining computational efficiency through PEFT via LoRA. The LLM judge introduces a potential source of noise in reward estimation, particularly for reasoning quality assessment where evaluation is inherently more subjective. Empirical validation in diverse model architectures establishes the robustness and generalizability of our methodology, with the Qwen3-VL-4B-Instruct model enhanced with reinforcement learning achieving superior accuracy (0.634) compared to larger foundation models while reducing computational overhead by 50% and inference latency by 71%. The framework's ability to achieve competitive performance with significantly reduced parameters and training costs positions it as a practical approach for real-world deployment, contributing to the broader goal of developing truly intelligent systems capable of sophisticated visual reasoning and logical inference.

Future work will focus on enhancing the reinforcement learning framework to further improve complex reasoning capabilities. Specifically, we intend to implement a multi-stage reward refinement, where initial RL training with an LLM judge is followed by stages incorporating human feedback or ensemble reward models. This will reduce potential reward hacking and enable the model to generalize better to expand the ceiling of achievable performance.

# A Full Response from All Models in the Examples

## A.1 Example 1: open-ended question

**Claude Sonnet 3.7**

<think>

To find the difference between the sum of the blue bars and the sum of the yellow bars, I need to:

1. Identify which bars are blue and which are yellow in the chart
2. Sum up all the blue bars
3. Sum up all the yellow bars
4. Calculate the difference between these sums

Looking at the chart:

- The blue bars appear to be labeled as “Right” in the legend
- The yellow bars appear to be labeled as “Centre” in the legend

Blue bars (Right) values:

- Kevin Rudd: 32
- Julia Gillard: 38
- Don’t know: 30

Sum of blue bars = 32 + 38 + 30 = 100

Yellow bars (Centre) values:

- Kevin Rudd: 55
- Julia Gillard: 27
- Don’t know: 12

Sum of yellow bars = 55 + 27 + 12 = 94

Difference between sum of blue bars and sum of yellow bars = 100 - 94 = 6

</think>

<answer>

The sum of the blue bars (Right) is 32 + 38 + 30 = 100.

The sum of the yellow bars (Centre) is 55 + 27 + 12 = 94.

The difference between the sum of the blue bars and the sum of the yellow bars is 100 - 94 = 6.

</answer>

**Qwen3-VL-4B-Instruct**

<answer>

12

</answer>

**Qwen3-VL-4B-Instruct-GRPO**

<think>

To calculate the difference between the sum of the blue bars and the sum of the yellow bars, I need to:

1. Identify all the blue bars and sum their values.
2. Identify all the yellow bars and sum their values.
3. Subtract the sum of the yellow bars from the sum of the blue bars.

From the chart:

- Blue bars (representing “Right”) have values: 32 (Kevin Rudd), 38 (Julia Gillard), and 30 (Don’t know).
- Yellow bars (representing “Centre”) have values: 55 (Kevin Rudd), 27 (Julia Gillard), and 18 (Don’t know).

Sum of blue bars = 32 + 38 + 30 = 100.

Sum of yellow bars = 55 + 27 + 18 = 100.

Difference = Sum of blue bars - Sum of yellow bars = 100 - 100 = 0. </think>

<answer>

0

</answer>

**Qwen3-VL-4B-Instruct-DAPO**

<think>

To calculate the difference between the sum of the blue bars and the sum of the yellow bars, I need to:

1. Identify all the blue bars and sum their values.
2. Identify all the yellow bars and sum their values.
3. Subtract the sum of the yellow bars from the sum of the blue bars.

From the chart:

- Blue bars (representing “Right”) have values: 32 (Kevin Rudd), 38 (Julia Gillard), and 30 (Don’t know).
- Yellow bars (representing “Centre”) have values: 55 (Kevin Rudd), 27 (Julia Gillard), and 18 (Don’t know).

Sum of blue bars = 32 + 38 + 30 = 100

Sum of yellow bars = 55 + 27 + 18 = 100

Difference = Sum of blue bars - Sum of yellow bars = 100 - 100 = 0

</think>

<answer>

0

</answer>

**Qwen3-VL-4B-Instruct-GSPO**

<think>

To calculate the difference between the sum of the blue bars and the sum of the yellow bars, we need to:

1. Identify the values of all blue bars (representing “Right”).
2. Identify the values of all yellow bars (representing “Centre”).
3. Sum the values of the blue bars.
4. Sum the values of the yellow bars.
5. Subtract the sum of the yellow bars from the sum of the blue bars.

From the chart:

Blue bars (Right):

- Kevin Rudd: 32
- Julia Gillard: 38
- Don’t know: 30
- Sum of blue bars = 32 + 38 + 30 = 100

Yellow bars (Centre):

- Kevin Rudd: 55

- Julia Gillard: 27
- Don't know: 18
- Sum of yellow bars = 55 + 27 + 18 = 100

Now, calculate the difference:

Difference = Sum of blue bars - Sum of yellow bars = 100 - 100 = 0

</think>

<answer>

0

</answer>

### A.2 Example 2: Multiple Choice

**Claude Sonnet 3.7**

<think>

To solve this problem, I need to:

1. Identify the pension fund assets values from 2010 to 2014
2. Calculate the growth trend during this period
3. Apply this trend to estimate the 2015 value
4. Compare the result with the given options

Looking at the chart, I can approximate the values:

- 2010: ~1,800 N billion
- 2011: ~2,400 N billion
- 2012: ~2,700 N billion
- 2013: ~3,800 N billion
- 2014: ~4,500 N billion

I can calculate the year-over-year growth rates:

- 2010 to 2011: (2400-1800)/1800 = 600/1800 = 0.33 or 33%
- 2011 to 2012: (2700-2400)/2400 = 300/2400 = 0.125 or 12.5%
- 2012 to 2013: (3800-2700)/2700 = 1100/2700 = 0.41 or 41%
- 2013 to 2014: (4500-3800)/3800 = 700/3800 = 0.18 or 18%

The growth rates vary significantly. I can calculate the average growth rate:

(33% + 12.5% + 41% + 18%)/4 = 104.5%/4 = 26.1%

Alternatively, I can calculate the compound annual growth rate (CAGR) from 2010 to 2014:

CAGR = (Final Value/Initial Value)$^{(1/\text{number of years})}$ - 1

CAGR = (4500/1800)$^{(1/4)}$ - 1

CAGR = (2.5)$^{0.25}$ - 1

CAGR = 1.26 - 1 = 0.26 or 26%

Using this growth rate to estimate 2015:

2015 estimate = 4500 × (1 + 0.26) = 4500 × 1.26 = 5670 N billion

This is closest to option c) 5,500 N billion.

However, I notice that the growth has been accelerating in the later years. If I just use the most recent growth rate (2013 to 2014) of 18%:

2015 estimate = 4500 × (1 + 0.18) = 4500 × 1.18 = 5310 N billion

This is closest to option b) 5,200 N billion.

Given the variability in growth rates and the fact that the question asks for the continuation of the trend from 2010 to 2014, I believe the CAGR approach is most appropriate, which gives us an estimate closest to option c) 5,500 N billion.

</think>

<answer>

Based on the chart showing pension fund assets from 2007-2014, I need to estimate the 2015 value by analyzing the growth trend from 2010-2014.

Looking at the approximate values from the chart:

- 2010: ~1,800 N billion
- 2011: ~2,400 N billion
- 2012: ~2,700 N billion
- 2013: ~3,800 N billion
- 2014: ~4,500 N billion

To find the overall growth trend, I'll calculate the compound annual growth rate (CAGR) from 2010 to 2014:

CAGR = (Final Value/Initial Value)$^{(1/\text{number of years})}$ - 1

CAGR = (4500/1800)$^{(1/4)}$ - 1

CAGR = (2.5)$^{0.25}$ - 1

CAGR = 1.26 - 1 = 0.26 or 26%

Applying this growth rate to estimate 2015:

2015 estimate = 4500 × (1 + 0.26) = 4500 × 1.26 = 5670 N billion

This is closest to option c) 5,500 N billion.

Therefore, the answer is c) 5,500 N billion.

</answer>

**Qwen3-VL-4B-Instruct**

<answer>

uncertain

</answer>

**Qwen3-VL-4B-Instruct-GRPO**

<think>

To estimate the pension fund assets in 2015 assuming the growth trend from 2010 to 2014 continues, we need to analyze the growth pattern in the bar chart.

**Step 1:** Observe the growth from 2010 to 2014.

- 2010: ~2,200 N billion
- 2011: ~2,500 N billion
- 2012: ~2,800 N billion
- 2013: ~3,800 N billion
- 2014: ~4,500 N billion

**Step 2:** Calculate the growth from 2013 to 2014.

- Growth = 4,500 - 3,800 = 700 N billion

**Step 3:** Assume the growth continues at a similar rate. The growth from 2013 to 2014 is 700 N billion. If we assume the same growth rate, then 2015 would be approximately 4,500 + 700 = 5,200 N billion.

**Step 4:** Compare with the options.

- a) 4,800 — too low
- b) 5,200 — matches our estimate
- c) 5,500 — higher than estimated
- d) 6,000 — too high

Therefore, the estimated pension fund assets in 2015 would be 5,200 N billion.

</think>
<answer>
b) 5,200
</answer>

**Qwen3-VL-4B-Instruct-DAPO**

<think>

To estimate the pension fund assets in 2015 assuming the growth trend from 2010 to 2014 continues, we need to analyze the growth pattern in the bar chart.

**Step 1:** Observe the growth from 2010 to 2014.

- 2010: ~2,200 N billion
- 2011: ~2,500 N billion
- 2012: ~2,800 N billion
- 2013: ~3,800 N billion
- 2014: ~4,500 N billion

**Step 2:** Calculate the growth from 2013 to 2014.

- Growth = 4,500 - 3,800 = 700 N billion

**Step 3:** Assume the growth continues at a similar rate. The growth from 2013 to 2014 is 700 N billion. If we assume the same growth rate, then 2015 would be approximately:

- 4,500 + 700 = 5,200 N billion

**Step 4:** Compare with the options.

- a) 4,800 — too low
- b) 5,200 — matches our estimate
- c) 5,500 — higher than estimated
- d) 6,000 — too high

Therefore, the estimated pension fund assets in 2015 would be 5,200 N billion.

</think>
<answer>
b) 5,200
</answer>

**Qwen3-VL-4B-Instruct-GSPO**

<think>

To estimate the pension fund assets in 2015 assuming the growth trend from 2010 to 2014 continues, we need to analyze the growth pattern in the bar chart.

**Step 1:** Observe the values from 2010 to 2014.

- 2010: ~2,200 N billion
- 2011: ~2,500 N billion
- 2012: ~2,800 N billion
- 2013: ~3,800 N billion
- 2014: ~4,500 N billion

**Step 2:** Calculate the growth from 2013 to 2014.

- Growth = 4,500 - 3,800 = 700 N billion

**Step 3:** Assume the growth continues at a similar rate. The growth from 2012 to 2013 was 1,000 N billion (3,800 - 2,800), and from 2011 to 2012 was 300 N billion (2,800 - 2,500). The growth is not linear, but the most recent growth (2013 to 2014) is 700 N billion.

**Step 4:** Estimate the growth from 2014 to 2015.

- If we assume the growth continues at the rate of the last year (700 N billion), then:
- 2015 ≈ 4,500 + 700 = 5,200 N billion

Alternatively, if we look at the overall trend from 2010 to 2014, the growth is accelerating. From 2013 to 2014, the growth was 700 N billion. If we assume a similar or slightly higher growth rate, 5,200 is a reasonable estimate.

Thus, the estimated pension fund assets in 2015 would be approximately 5,200 N billion.

</think>
<answer>
b) 5,200
</answer>

### A.3 Example 3: True/False Question

**Claude Sonnet 3.7**

<think>

Let me analyze the chart in the top left corner titled "OPEN CASES: BY CREATED DATE" to determine if there is a positive correlation between account value and creation date.

The chart shows:

1. X-axis: Case Creation Date (ranging from approximately April to August)
2. Y-axis: Account Value (ranging from $0 to approximately $150K)
3. Bubbles of different sizes and colors (green and red) representing open cases

To determine if there's a positive correlation, I need to see if there's a trend where account values increase as the creation date becomes more recent.

Looking at the scatter plot:

- The data points are scattered across the entire date range
- There are high-value accounts (around $100K-$150K) distributed throughout the time period
- There are low-value accounts distributed throughout the time period
- There doesn't appear to be a clear upward trend from left to right

Based on visual inspection, I don't see a clear positive correlation between account value and creation date. The high-value accounts appear to be distributed across different dates rather than concentrated in more recent dates. If anything, the distribution appears relatively random.

</think>
<answer>

The statement is incorrect. The chart in the top left corner titled “OPEN CASES: BY CREATED DATE” does not show a positive correlation between account value and creation date.

Looking at the scatter plot, the account values (Y-axis ranging from $0 to approximately $150K) appear to be distributed relatively randomly across the entire date range (X-axis from April to August). High-value accounts (around $100K-$150K) and low-value accounts are scattered throughout the time period without showing a clear upward trend from earlier to later dates.

If there were a positive correlation, we would expect to see account values consistently increasing as we move from left (earlier dates) to right (later dates) on the chart, which is not evident in this visualization.

</answer>

**Qwen3-VL-4B-Instruct**

<answer>

uncertain

</answer>

**Qwen3-VL-4B-Instruct-GRPO**

<think>

To evaluate whether the chart in the top-left corner indicates a positive correlation between account value and case creation date, I need to analyze the scatter plot.

**Step 1:** Understand the chart’s axes.

- The x-axis represents “Case Creation Date” (with dates like 7-Apr, 28-Apr, etc.).
- The y-axis represents “Account Value” (in $K, ranging from $0K to $200K).

**Step 2:** Observe the pattern of the data points.

- The green circles (larger bubbles) are clustered toward the right side of the chart (later dates) and also at higher account values (e.g., $100K, $200K).
- The red circles (smaller bubbles) are clustered toward the left side (earlier dates) and at lower account values (e.g., $50K, $100K).

**Step 3:** Determine if there’s a positive correlation.

- A positive correlation would mean that as the case creation date moves forward (to the right), account value also increases.
- In this chart, the larger bubbles (representing higher account values) are indeed located toward the right (later dates), while smaller bubbles (lower account values) are on the left (earlier dates).

**Step 4:** Conclusion.

- The visual pattern suggests that cases created later (later dates) tend to have higher account values, which indicates a positive correlation.

<answer>

True

</answer>

**Qwen3-VL-4B-Instruct-DAPO**

<think>

To evaluate whether the chart in the top-left corner indicates a positive correlation between account value and case creation date, I need to analyze the scatter plot.

**Step 1:** Understand the chart.

- The x-axis represents “Case Creation Date” (with dates like 7-Apr, 28-Apr, etc.).
- The y-axis represents “Account Value” (in $K, so $100K, $200K, etc.).
- The dots represent individual open cases, with size possibly indicating case volume or importance.

**Step 2:** Look for a trend.

- A positive correlation would mean that as the case creation date moves forward (to the right on the x-axis), the account value (y-axis) also increases.
- I need to visually inspect if dots tend to cluster in a way that higher account values are associated with later dates.

**Step 3:** Analyze the visual pattern.

- The chart shows a cluster of dots on the left (earlier dates) with relatively lower account values (e.g., around $50K).
- As we move to the right (later dates), there are dots with higher account values (e.g., up to $200K).
- This suggests that cases created later tend to have higher account values.

**Step 4:** Conclusion.

- The visual trend supports a positive correlation: later case creation dates are associated with higher account values.

</think>

<answer>

True

</answer>

**Qwen3-VL-4B-Instruct-GSPO**

<think>

To evaluate whether the chart in the top-left corner indicates a positive correlation between account value and case creation date, let’s break this down step by step:

**1. Understand the chart:** The top-left chart is titled “OPEN CASES • BY CREATED DATE.” It plots “Account Value” (y-axis) against “Case Creation Date” (x-axis). The data points are represented by circles, with size possibly indicating case volume or importance.

**2. What does “positive correlation” mean?** A positive correlation implies that as one variable increases, the other variable also tends to increase. In this case, if account value increases as case creation date moves forward (e.g., from early to late months), then there’s a positive correlation.

**3. Analyze the chart visually:**

- The x-axis represents case creation dates, spanning from early April to late June.
- The y-axis represents account value, ranging from $0K to $200K.
- The data points (circles) are clustered in different regions. Notice that the largest circles (representing higher account values) appear toward the right side of the chart (later dates), while smaller circles (lower account values) are more clustered on the left (earlier dates).

**4. Interpretation:**

- The trend suggests that cases created later in the period (right side of the x-axis) tend to have higher account values (top of the y-axis).
- This indicates that as the case creation date progresses, account value tends to increase — which is consistent with a positive correlation.

**5. Conclusion:**

- The visual pattern supports the claim of a positive correlation between account value and case creation date.

```
</think>
<answer>
True
</answer>
```